\documentclass{article}

\usepackage{arxiv}

\usepackage[utf8]{inputenc} 
\usepackage[T1]{fontenc}    
\usepackage{hyperref}       
\usepackage{url}            
\usepackage{booktabs}       
\usepackage{amsfonts}       
\usepackage{nicefrac}       
\usepackage{microtype}      
\usepackage{lipsum}		
\usepackage{amsmath}
\usepackage{cleveref} 
\usepackage{graphicx}
\usepackage{natbib}
\usepackage{doi}
\usepackage{amsmath}

\usepackage{algorithm}
\usepackage{algorithmic}
\usepackage{multirow}
\usepackage[table]{xcolor}
\usepackage[section]{placeins}

\title{Adv-TGD: Adversarial Text-Guided Diffusion for Face Recognition Impersonation Attacks}

\author{{\includegraphics[scale=0.06]{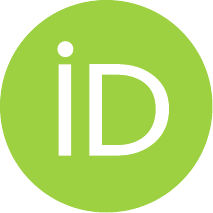}\hspace{1mm}Omid Ahmadieh} \\
		Bellini College of Artificial Intelligence\\
        Cybersecurity and Computing\\
	University of South Florida\\
	Tampa, FL\\\\
	\And
	\href{https://orcid.org/0000-0002-4590-7170}{\includegraphics[scale=0.06]{orcid.pdf}\hspace{1mm}Nima Karimian} \\
	Bellini College of Artificial Intelligence,\\ Cybersecurity and Computing\\
	University of South Florida\\
	Tampa, FL\\
}

\renewcommand{\shorttitle}{\textit{arXiv} Template}

\title{Adv-TGD: Adversarial Text-Guided Diffusion for Face Recognition Impersonation Attacks}

\begin{document}
\maketitle

\begin{abstract}
The widespread adoption of face recognition (FR) technologies raises serious privacy concerns, as facial data can be exploited without consent. To address this challenge, we propose \textbf{Adv-TGD}, a generative adversarial attack framework that synthesizes photorealistic faces capable of impersonating target identities and deceiving face recognition systems. Built upon Stable Diffusion v2.1, Adv-TGD performs per-sample LoRA fine-tuning conditioned on concise textual prompts to generate natural yet adversarially manipulated identities. Unlike conventional identity-attack approaches, our method optimizes lightweight cross-attention adapters for each source–target pair within a fixed-timestep denoising process. Latent blending is constrained by a face-local heatmap mask to ensure spatially precise identity manipulation while preserving non-sensitive regions. We introduce a composite objective that integrates masked $\epsilon$-MSE reconstruction, thresholded identity divergence in FR embedding space, directional feature alignment, and source-similarity suppression to balance adversarial attack and visual realism. Optionally, LLaVA-generated attribute prompts enhance fine-grained semantic details without reintroducing identity cues. Under the black-box evaluation protocol, \textbf{Adv-TGD} attains an average attack success rate (ASR) of \textbf{85.90\%} across IR152, IRSE50, MobileFace, and FaceNet surpassing the semantic SOTA baseline Adv-CPG by \textbf{+6.25} points, diffusion-based makeup method DiffAIM by \textbf{+3} points, and noise-based P3-Mask by \textbf{+16} points. Despite its strong attack efficacy, Adv-TGD preserves high visual fidelity (\textbf{PSNR = 28.18 dB}, \textbf{SSIM = 0.981}). Furthermore, we demonstrate the flexibility of our framework by successfully extending it to in-the-wild datasets (LADN), general object classification (ImageNet), and transformer-based diffusion models (FLUX.1). These results demonstrate that text-guided, per-pair LoRA adaptation offers a practical and effective paradigm for adversarial face manipulation for impersonation attacks, unifying semantic controllability, photorealism, and robustness against unauthorized face recognition.

\end{abstract}


\section{Introduction}
\label{sec:intro}

\begin{figure*}[t]
  \centering
  \includegraphics[width=0.95\textwidth]{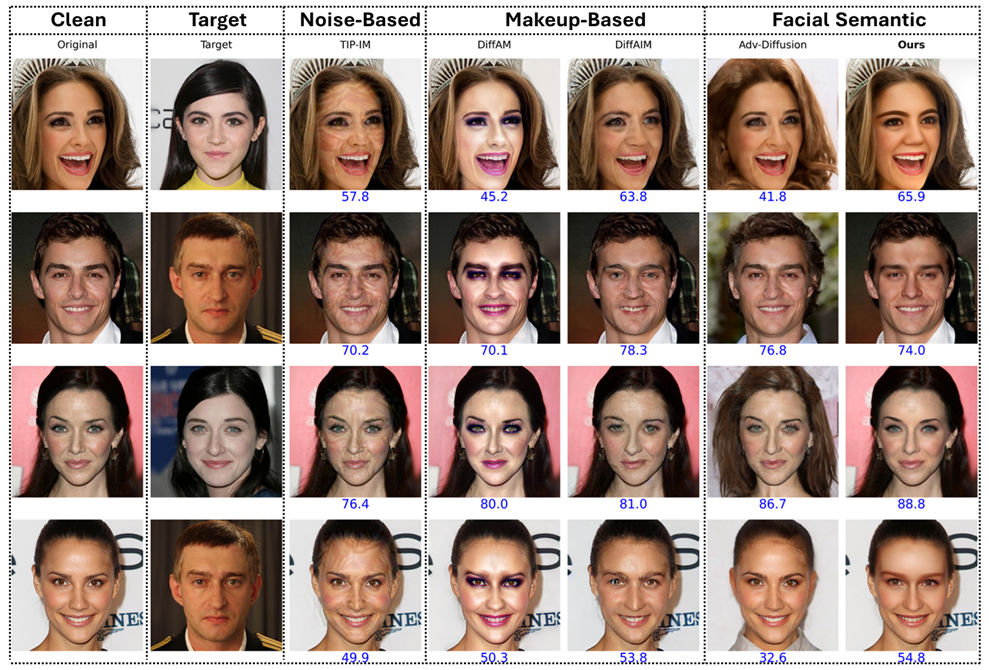}
  \caption{The proposed \textbf{Adv-TGD} achieves realistic, identity-aligned face transformations while preserving texture, expression, and lighting consistency. The blue number above each image denotes the \textit{Face++} confidence score.}
  \label{fig:teaser}
\end{figure*}

The rapid advancement of deep neural networks has propelled face recognition (FR) systems to remarkable success across applications such as authentication, surveillance, and social media tagging~\cite{mi2024privacy, pinto2011scaling}. However, their widespread deployment has raised serious privacy concerns. By exploiting large-scale public image datasets, FR models can infer social relationships~\cite{zhang2018facial}, facilitate identity theft~\cite{wang2025nullswap,wenger2023sok}, and even enable unauthorized mass surveillance~\cite{almeida2022ethics}. Despite their impressive accuracy, modern face recognition systems remain vulnerable to adversarial manipulations. Understanding these vulnerabilities is critical for improving the robustness and security of biometric systems. In this work, we propose Adv-TGD, a diffusion-based adversarial attack framework that generates photorealistic facial images capable of impersonating target identities and deceiving multiple face recognition models.

To counter these threats, adversarial attacks have emerged as a powerful tool. Early methods applied adversarial perturbations~\cite{fan2020sparse,mirjalili2020privacynet,chhabra2025pridentity,wang2023privacy,wu2018towards} to deceive FR models, but traditional noise-based~\cite{madry2018towards,yang2021towards} and patch-based~\cite{wei2025distributionally,wei2025real,ma2023transferable,wang2022patchnet,feng2023dynamic,wei2022adversarial,wei2022simultaneously,yang2023towards} attacks, including pioneering physical accessories like adversarial eyeglasses~\cite{sharif2016accessorize}, often produce visible artifacts, making them unsuitable for natural image sharing. A practical attack framework should protect identity while preserving photorealism. GAN-based editing~\cite{liu2023gan,huang2020pfa} has recently improved realism via makeup transfer and semantic manipulation. These concepts have been extended to physical style impersonations~\cite{an2023imu}, but it requires target-specific retraining and offer limited generalization within the GAN manifold.
Diffusion models~\cite{sun2024diffam,liu2024adv,wang2025adv,liu2023diffprotect,kim2023dcface,stypulkowski2024diffused} overcome these limits, providing high realism and controllability. Recent works--DiffProtect~\cite{liu2023diffprotect}, Adv-Diffusion~\cite{liu2024adv}, and Adv-CPG~\cite{wang2025adv}--pioneered diffusion-based adversarial generation to protect facial privacy through latent perturbation and text-conditioned generation. However, they still rely on iterative optimization and target-specific fine-tuning, limiting efficiency and editability.

\begin{figure*}[t!]
  \centering
  \includegraphics[width=0.95\linewidth]{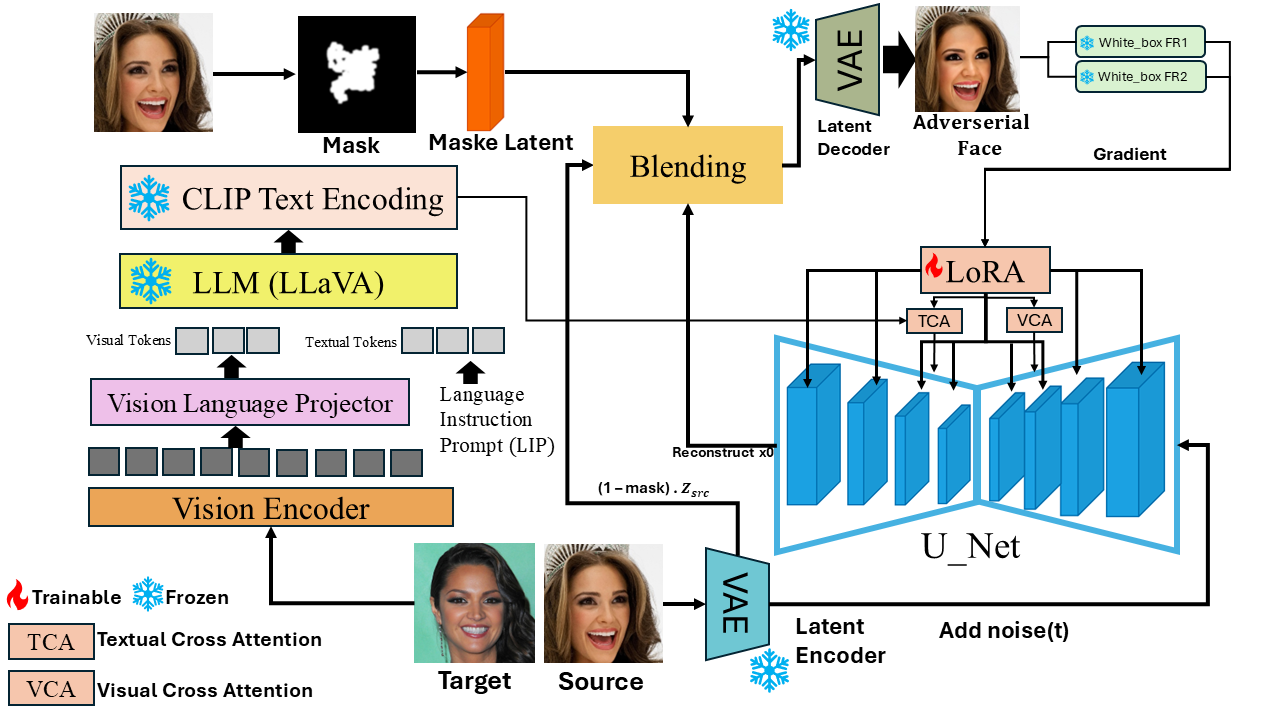}
  \caption{\textbf{Architecture of \textit{Adv-TGD}.}
  Per-pair LoRA fine-tuning on a frozen SD 2.1 U-Net utilizing a single-pass latent reconstruction objective. A face-local SGSM produces a latent-resolution gate for masked latent blending; decoded images feed identity, directional, source-suppression, and late masked text losses.
  Re-evaluation blends the top-scoring aligned frame back into the original photo and reports ASR/PSNR/SSIM.}
  \label{fig:arch}
\end{figure*}

To address these limitations, we propose \textbf{Adv-TGD}, a text-conditioned adversarial generation framework built on Stable Diffusion 2.1 (Fig.~\ref{fig:arch}). Our model performs per-pair LoRA adaptation within the U-Net's cross-attention layers, enabling single-pass latent reconstruction during the 65-iteration optimization for each source–target identity pair while keeping the VAE, text encoder, and U-Net backbone frozen. As shown in Fig.~\ref{fig:arch}, given a source image and a target identity, we utilize a hybrid \textbf{Salience-Guided Semantic Mask (SGSM)} that fuses anatomical landmarks with aggregated FR gradients to localize identity-critical regions.

Within these regions, lightweight adapters are optimized to inject adversarial semantics guided by LLaVA-generated language prompts. Unlike traditional noise-based attacks, our framework targets low-to-mid frequency structural features, successfully neutralizing recognizers by dispersing their spatial attention a process we quantify via spatial spread and entropy metrics. Text guidance is applied late and locally as a lightweight regularizer for attribute consistency (e.g., hair or accessories) rather than a primary driver of identity shift. Training uses a fixed diffusion step and \textit{latent blending} so that FR models ``see'' the edited face on the original background; at inference, we invert the alignment and use seamless cloning (Normal Clone) to reinsert the result into the source frame, ensuring photorealistic reconstruction without edge artifacts or identity leakage.

In summary, our main contributions are:
\begin{itemize}
\item \textbf{Text-conditioned adversarial diffusion.} We present a diffusion-based impersonation framework that leverages natural-language guidance and lightweight LoRA adapters to generate adversarially shifted yet visually coherent identities. Our framework generate photorealistic impersonation images capable of deceiving multiple face recognition systems.

\item \textbf{Automatic semantic prompt generation via vision language models.}
We leverage LLaVA to automatically generate descriptive language prompts from the target image, enabling fine-grained semantic guidance for diffusion-based identity manipulation. This removes the need for manual prompt engineering while improving attribute consistency during adversarial face generation.

\item \textbf{Salience-Guided Semantic Masking (SGSM).} We introduce a hybrid masking strategy that integrates anatomical landmark priors with ensemble gradient saliency. By enforcing precise $top$-$k$ spatial support within a semantic facial hull, SGSM selectively amplifies identity-critical regions and yields a contiguous, soft latent-space mask, achieving accurate identity localization without manual annotation.

\item \textbf{Unified adversarial objective.} We develop a principled composite loss that integrates masked denoising, a threshold-aware softmin identity hinge, directional $source \to target$ alignment, and source-suppression regularization. This formulation is designed to maximize spatial attention dispersion, forcing the recognizer to search a wider, non-identity-specific radius.

\item \textbf{Cross-architecture attack generalization.} 
We show that Adv-TGD is architecture-agnostic and can be applied to both UNet-based diffusion models (Stable Diffusion 1.5 and 2.1) and transformer-based generative models (FLUX.1). Extensive experiments demonstrate that the proposed attack maintains strong impersonation success across different generative backbones, revealing a broader vulnerability of modern generative diffusion systems.

\item \textbf{Realism and robustness.} We introduce a practical post-processing pipeline combining inverse alignment and seamless blending to restore natural context and eliminate boundary artifacts. The proposed approach preserves high visual fidelity (PSNR = 28.18 dB, SSIM = 0.981) while maintaining strong adversarial effectiveness, achieving state-of-the-art impersonation performance with an average ASR of 85.90\% across multiple face recognition models (IR152, IRSE50, MobileFace, and FaceNet).
\end{itemize}

\section{Related Work}
\label{sec:related_work}

\subsubsection{Diffusion Models for Semantic Image Manipulation}
Diffusion models~\cite{ho2020denoising,song2020score,dhariwal2021diffusion,rombach2022ldm} have revolutionized generative modeling by iteratively refining noise into coherent imagery. Latent formulations like Stable Diffusion~\cite{rombach2022ldm,podell2023sdxl} enable semantically guided manipulation under text, mask, or structural control.
Parameter-efficient adapters such as LoRA~\cite{hu2022lora} and T2I-Adapters~\cite{mou2023t2iadapter} add flexibility through low-rank updates without retraining the backbone. These properties make diffusion priors well-suited for adversarial face generation, where both realism and identity consistency are essential.

\subsubsection{Text-Conditioned Personalization and Identity Adaptation}
Text-conditioned personalization frameworks aim to retain individual identity while providing stylistic or semantic control through language~\cite{balaji2022ediff, Composer, nichol2021glide, ramesh2022hierarchical, saharia2022photorealistic, rombach2022high}.
Recent approaches leverage adapter-based or prompt-tuning strategies~\cite{hu2022lora, kumari2023multi} to achieve rapid adaptation without retraining the entire model. However, most focus on visual customization rather than adversarial transformation. Our work revisits this paradigm, integrating LoRA-based textual conditioning with adversarial identity guidance to achieve semantically meaningful, stealthy adversarial impersonation. Given a target face $I_{tgt}$, LLaVA produces a textual description capturing high-level attributes such as hairstyle, facial structure, accessories, and appearance characteristics. The generated description is converted into a concise natural-language prompt $c$, which conditions the diffusion model during optimization. Unlike manual prompt engineering used in many text-guided diffusion methods, this automatic prompt generation ensures consistent semantic guidance while minimizing human intervention.

\subsubsection{Adversarial Attacks for Face Recognition}
Adversarial perturbations are widely used to bypass and evaluate unauthorized FR systems. Existing methods fall into \emph{noise-based} and \emph{unrestricted} categories. Noise-based approaches~\cite{cherepanova2021lowkey,shan2020fawkes,yang2021towards,dong2018boosting,dong2019evading,madry2018towards} inject $\ell_p$-bounded pixel noise or apply data poisoning and game-theoretic strategies to reduce recognition accuracy, but they often require model gradients or multiple samples and produce visible artifacts. Black-box variants like TIP-IM~\cite{yang2021towards} relax these constraints yet still compromise realism. Unrestricted attacks embed perturbations in semantic edits~\cite{yin2021adv}, improving stealth but showing poor transferability and unnatural textures on unseen FR models.

\subsubsection{Makeup-Based and Generative Adversarial Methods}
Recent studies leverage facial makeup as a natural adversarial medium to conceal identity cues~\cite{komkov2021advhat,yin2021adv,hu2022protecting,shamshad2024makeup,zeng,zhu2019generating}.
By embedding adversarial signals within realistic cosmetic edits, these methods achieve high perceptual quality while preserving attack efficacy.
However, reliance on large-scale makeup datasets introduces demographic bias and limits personalization when source and reference domains differ. Text-guided frameworks such as CLIP2Protect~\cite{shamshad2023clip2protect} mitigate these issues via semantic control but still inherit StyleGAN-related biases and limited cultural diversity~\cite{munoz2023uncovering,karakas2022fairstyle,yu2025data,perera2025unbiased}. Moreover, most approaches rely on iterative optimization and lack strong generative priors for structure preservation; despite improved transferability from unrestricted and diffusion-based attacks, achieving consistently high ASR (proportion of generated images that are verified as the target identity by the recognition system) across models remains difficult.

\subsubsection{Diffusion-Driven Adversarial Generation}
Integrating diffusion priors into adversarial attack design has emerged as a strong alternative to pixel-space optimization.
Adv-Diffusion~\cite{liu2024adv} introduced latent perturbations but required multi-step optimization and lacked semantic control.
Later variants such as Diffam~\cite{sun2024diffam} improved photorealism yet struggled to balance fidelity and adversarial strength.
Our \emph{Adv-TGD} advances this line by embedding text-guided, LoRA-based adaptation within Stable Diffusion~2.1's cross-attention layers.
It performs single-pass latent reconstruction during per-pair optimization with FR-aware masking and late textual regularization, achieving state-of-the-art ASR (Table~\ref{tab:asr_results_columnar}) while maintaining contextual and identity realism.
Motivated by the limitations of iterative or visually intrusive attacks, \emph{Adv-TGD} serves as a unified, text-conditioned diffusion framework enabling fast, localized, and photorealistic identity impersonation.

\section{Methodology}
\label{sec:methodology}

We introduce \textbf{Adv-TGD}, a per-sample LoRA fine-tuning framework that synthesizes \emph{targeted} adversarial faces by adapting a frozen latent diffusion model with a lightweight, pair-specific adapter. The method has five ingredients: (i) a \emph{face-local SGSM} that localizes edits to identity-bearing facial regions, (ii) \emph{ensemble identity guidance} with a threshold-aware hinge aggregated by a softmin to remove weakest-view failure modes, (iii) a \emph{directional identity} term that encourages movement toward the target in embedding space, (iv) \emph{source suppression} to avoid snapping back to the source identity, and (v) a \emph{late, masked text--image} loss to maintain attribute consistency. We train on aligned faces and seamlessly warp the result back into the original frame.

\subsection{Preliminaries}
\subsubsection{Latent diffusion}
We build on Stable Diffusion v2.1~\cite{rombach2022high} with encoder--decoder $(\mathcal{E},\mathcal{D})$ and denoiser $\epsilon_\theta$ acting on latents. For an image $I$, $z=\mathcal{E}(I)$ and $\hat{I}=\mathcal{D}(z)$. At noise index $t$ with cumulative $\alpha_t\in(0,1]$, we corrupt $z$ to:
\begin{displaymath}
z_t=\alpha_t^{1/2}z+(1-\alpha_t)^{1/2}\epsilon,\quad \epsilon\sim\mathcal{N}(0,I),
\end{displaymath}
and predict $\hat{\epsilon}_\theta(z_t,t,c)$ under text condition $c$.

\subsubsection{Parameter-efficient adaptation}
For each source--target pair, we insert LoRA adapters into the U-Net cross-attention projections (\texttt{to\_q}, \texttt{to\_k}, \texttt{to\_v}, \texttt{to\_out.0}) and optimize only those parameters. The VAE, text encoder, and the base U-Net remain frozen.

\subsection{Overview}
\label{sec:overview}

The fundamental challenge in practical adversarial impersonation is balancing identity manipulation (deceiving the face recognition model) with visual stealth (preserving photorealism and context). To achieve this, our attack methodology is built upon three strategic design choices. First, rather than applying global pixel noise that degrades image quality, we ensure localized semantic manipulation by restricting adversarial edits strictly to identity-bearing facial structures. This is achieved via a hybrid mask that targets the specific spatial regions face recognition models rely on, bounded by anatomical priors. Second, to bypass computationally heavy iterative diffusion inversion, we perform parameter-efficient optimization via single-pass latent reconstruction. By freezing the base diffusion model and optimizing lightweight LoRA adapters under a fixed-timestep denoising objective, we efficiently execute structural identity changes without requiring full diffusion trajectories. Finally, to ensure successful impersonation without drifting into unnatural facial distributions, we employ multi-objective identity guidance. This composite loss landscape simultaneously pushes the generated identity toward the target, enforces directional semantic consistency, suppresses the source identity, and preserves visual attributes via text priors.

Practically, our pipeline executes these principles in three consecutive stages. Given a source $I_{\text{src}}$ and a target $I_{\text{tgt}}$, we (1) align both faces and construct a soft, identity-aware mask $M$; (2) fine-tune a pair-specific LoRA via the single-step denoiser objective with latent blending guided by $M^\ell$ (the latent-resolution mask); and (3) invert the alignment and apply seamless blending to insert the edited face back into the original photo.

\subsection{Salience-Guided Semantic Masking (SGSM)}
\label{sec:sgsm}

To ensure that adversarial perturbations are both effective and spatially constrained, we propose a hybrid masking strategy that fuses anatomical priors with model-specific sensitivity. Let $\{\mathcal{F}_k\}_{k=1}^{K}$ denote an ensemble of FR models.

On the aligned source image $\mathbf{x}_s$, we first define a \emph{semantic prior} $M_{sem}$ as a landmark-based convex hull covering the eyes, nose, mouth, and jawline. Simultaneously, we derive a \emph{target-aware saliency map} $S$ by backpropagating the cosine similarity toward the target embedding:
\begin{equation}
    S = \frac{1}{K} \sum_{k=1}^K \left| \nabla_{\mathbf{x}_s} \cos(\mathcal{F}_k(\mathbf{x}_s), e^k_{tgt}) \right|.
\end{equation}
The raw saliency is gated by an elliptical face prior $P \in [0,1]^{H \times W}$ and normalized. To isolate the most identity-critical hotspots, we apply a thresholding operation, $\operatorname{TopK}(S, p\%)$, which retains only the top $p\%$ of pixels with the highest gradient saliency values. The final hybrid mask $M$ is defined as the union of the anatomical semantic prior $M_{\text{sem}}$ and these top salient regions, followed by a smoothing function:
\begin{equation}
    M = \operatorname{Feather}\left( M_{\text{sem}} \cup \operatorname{TopK}(S, p\%) \right),
\end{equation}
where $\operatorname{Feather}(\cdot)$ denotes a two-step post-processing operation comprising morphological dilation to slightly expand the mask boundaries, followed by Gaussian feathering to soften the edges. This hybrid approach ensures the optimizer aggressively targets specific identity-bearing structures (via $S$) while remaining strictly bounded by facial anatomy (via $M_{\text{sem}}$). Furthermore, the soft contiguous edges generated by the feathering operation prevent the background artifacts or ``ghosting'' often seen in unconstrained adversarial edits. Finally, the high-resolution mask is downsampled to $M^{\ell} \in [0,1]^{H/8 \times W/8}$ to guide the latent-space blending.

\begin{figure}[t]
    \centering
    \includegraphics[width=0.80\linewidth]{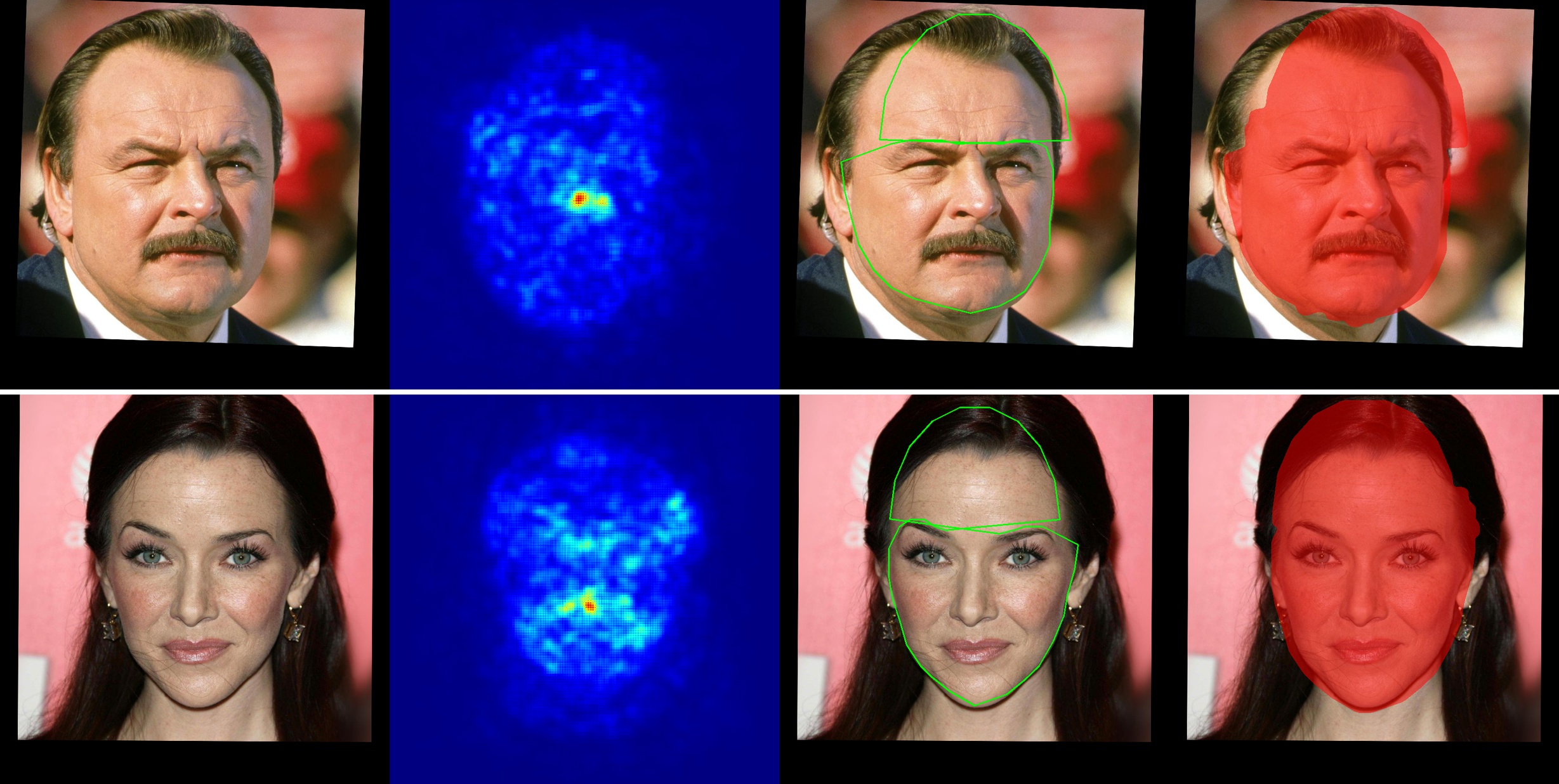}
    \caption{\textbf{Salience-Guided Semantic Masking (SGSM).} 
    Comparison across two identities showing (Col 1) Source, (Col 2) Saliency hotspots $S$, (Col 3) Semantic hull $M_{sem}$, and (Col 4) the final hybrid mask $M$. The strategy ensures targeted identity manipulation while maintaining anatomical grounding.}
    \label{fig:mask_showcase_single}
\end{figure}

\subsection{Fixed-Timestep Diffusion Denoising with LoRA Optimization}
\label{sec:trainstep}

At each iteration, we form a single noisy latent slice from the aligned source, predict noise with the LoRA-augmented UNet, reconstruct a clean latent, blend it with the source using a face mask, and decode to the image; only the LoRA weights are updated.

\vspace{1mm}
\noindent\textit{Form a noisy latent and predict noise:}
\begin{align}
z_{\text{src}} &= \mathcal{E}(I_{\text{src}}), \qquad \epsilon \sim \mathcal{N}(0,I), \\
z_t &= \sqrt{\alpha_t}\, z_{\text{src}} \;+\; \sqrt{1-\alpha_t}\, \epsilon, \\
\hat{\epsilon}_\theta &= \text{UNet}_{\text{LoRA}}(z_t,\, t,\, c).
\end{align}

\noindent \textit{where} $\mathcal{E}$ is the VAE encoder; $I_{\text{src}}$ is the aligned source face; $t = \lfloor \tau t_{\text{max}} \rfloor$ is the discrete timestep derived from a fixed diffusion index fraction $\tau \in (0,1)$, with $t_{\text{max}}$ representing the maximum number of diffusion timesteps; $\alpha_t$ is the scheduler's cumulative noise factor; $c$ is the textual conditioning; and $\text{UNet}_{\text{LoRA}}$ is the frozen UNet backbone equipped with trainable LoRA adapters in its cross-attention layers.

\noindent\textit{Reconstruct the clean latent according to the scheduler type:}

\begin{align}
\text{$\epsilon$-pred:}\quad \hat{z}_0 &= \big(z_t - (1-\alpha_t)^{1/2}\, \hat{\epsilon}_\theta\big)\, /\, \alpha_t^{1/2}, \label{eq:recon_eps}\\
\text{$v$-pred:}\quad \hat{z}_0 &= \alpha_t^{1/2}\, z_t \;-\; (1-\alpha_t)^{1/2}\, \hat{\epsilon}_\theta. \label{eq:recon_v}
\end{align}

\noindent\textit{where} the chosen case matches the diffusion scheduler's prediction parameterization.

\noindent\textit{Blend only the facial region in latent space and decode:}

\begin{align}
\hat{z}_0^{\text{blend}} &= M^\ell \odot \hat{z}_0 \;+\; (1-M^\ell) \odot z_{\text{src}}, \\
\hat{I} &= \mathcal{D}(\hat{z}_0^{\text{blend}}).
\end{align}

\noindent\textit{where} $M^\ell \in [0,1]^{H^\ell\times W^\ell}$ is the latent-resolution face mask; 
$\odot$ is element-wise multiply; $\mathcal{D}$ is the VAE decoder; 
$\hat{I}$ is used by the losses in \cref{sec:losses}.

\noindent\textit{Update step (LoRA only):}

\begin{equation}
\theta_{\text{LoRA}} \leftarrow \theta_{\text{LoRA}} - \eta\, \nabla_{\theta_{\text{LoRA}}}\,\mathcal{L},
\end{equation}

\noindent\textit{where} $\theta_{\text{LoRA}}$ denotes the trainable LoRA parameters, and $\eta$ is the learning rate. The VAE ($\mathcal{E}, \mathcal{D}$), text encoder, and base UNet remain completely frozen.

\begin{algorithm}[t]
\caption{Fixed-timestep LoRA fine-tuning (per source-target pair)}
\label{alg:train}
\begin{algorithmic}[1]
\REQUIRE Aligned $(I_{\text{src}}, I_{\text{tgt}})$, text $c$, masks $M$, $M^\ell$, FR ensemble $\{\mathcal{F}_k\}$, fixed index $t$
\ENSURE Optimized LoRA parameters $\theta_{\text{LoRA}}$ and adversarial image $\hat{I}$

\vspace{1mm} 
\STATE \COMMENT{\textbf{Step 1: Initialization \& Forward Diffusion}}
\STATE $z_{\text{src}} \gets \mathcal{E}(I_{\text{src}})$ \hfill \COMMENT{Encode source image to latent space}
\STATE Sample noise $\epsilon \sim \mathcal{N}(0,I)$
\STATE $z_t \gets \sqrt{\alpha_t} z_{\text{src}} + \sqrt{1-\alpha_t}\epsilon$ \hfill \COMMENT{Corrupt latent at fixed timestep $t$}

\vspace{1mm}
\STATE \COMMENT{\textbf{Step 2: Denoising \& Reconstruction}}
\STATE $\hat{\epsilon}_\theta \gets \text{UNet}_{\text{LoRA}}(z_t, t, c)$ \hfill \COMMENT{Predict noise using active LoRA + text prior}
\STATE Reconstruct clean latent $\hat{z}_0$ from $(z_t, \hat{\epsilon}_\theta)$ using Eqs.~\eqref{eq:recon_eps},\eqref{eq:recon_v}

\vspace{1mm}
\STATE \COMMENT{\textbf{Step 3: Masked Blending \& Decoding}}
\STATE $\hat{z}_0^{\text{blend}} \gets M^\ell \odot \hat{z}_0 + (1-M^\ell)\odot z_{\text{src}}$ \hfill \COMMENT{Restrict semantic edits to facial region}
\STATE $\hat{I} \gets \mathcal{D}(\hat{z}_0^{\text{blend}})$ \hfill \COMMENT{Decode blended latent back to pixel space}

\vspace{1mm}
\STATE \COMMENT{\textbf{Step 4: Optimization}}
\STATE Compute composite loss $\mathcal{L}$ comprising $\mathcal{L}_\epsilon, \mathcal{L}_{\text{id}}, \overline{\mathcal{D}}, \overline{\mathcal{S}}, \mathcal{L}_{\text{txt}}$
\STATE $\theta_{\text{LoRA}} \gets \theta_{\text{LoRA}} - \eta \nabla_{\theta_{\text{LoRA}}} \mathcal{L}$ \hfill \COMMENT{Gradient update on LoRA weights only}
\end{algorithmic}
\end{algorithm}

\subsection{Losses and Regularizers}
\label{sec:losses}

\subsubsection{FR Embedding Notation and Cosine Operators}
Before introducing the losses, we fix notation and define the quantities used throughout. We use the cosine similarity and the unit-normalization operator:
\begin{equation}
\label{eq:cos_unit}
\cos(u,v)=\frac{\langle u,v\rangle}{\|u\|\,\|v\|},
\qquad
\nu(x)=\frac{x}{\|x\|}.
\end{equation}
Given an ensemble of $K$ face recognizers $\{\mathcal{F}_k\}_{k=1}^K$, we denote the \emph{unit-normalized} embeddings of the (aligned) source image $I_{\text{src}}$, target image $I_{\text{tgt}}$, and the current prediction $\hat{I}$ as:
\begin{equation}
\label{eq:fr_embeds}
e^k_{\text{src}}=\nu\!\big(\mathcal{F}_k(I_{\text{src}})\big),\quad
e^k_{\text{tgt}}=\nu\!\big(\mathcal{F}_k(I_{\text{tgt}})\big),\quad
f_k(\hat{I})=\nu\!\big(\mathcal{F}_k(\hat{I})\big).
\end{equation}
All identity terms below are computed with these normalized embeddings.

\subsubsection{Training objective}
We combine a masked denoising term with identity guidance, directional consistency, source suppression, and a late text prior. The identity-related terms are smoothly ramped during training, while text guidance activates in later iterations:
\begin{equation}
\label{eq:tot}
\mathcal{L} = \lambda_{\epsilon}\,\mathcal{L}_{\epsilon} + s(i)\!\big(\lambda_{\text{id}}\mathcal{L}_{\text{id}} + \lambda_{\text{dir}}\overline{\mathcal{D}} + \lambda_{\text{src}}\overline{\mathcal{S}}\big) + \lambda_{\text{txt}}(i)\,\mathcal{L}_{\text{txt}}.
\end{equation}
\noindent\textit{where} $\lambda_\epsilon,\lambda_{\text{id}},\lambda_{\text{dir}},\lambda_{\text{src}}$ are fixed weights, 
$s(i)=b+(1-b)(i/N)^\gamma$ ramps identity terms over iteration $i$, and $\lambda_{\text{txt}}(i)$ turns on the text prior in the late phase.

\subsubsection{Latent-Space Masked Denoising Loss}
We restrict the denoising objective to the facial latent region, so capacity is focused where identity resides:
\begin{equation}
\label{eq:masked_mse}
\mathcal{L}_{\epsilon} = \frac{1}{Z}\sum_{c,i,j} M^\ell_{ij}\,\big(\epsilon_{cij}-\hat{\epsilon}_{\theta,cij}\big)^2,
\qquad
Z=C\sum_{i,j} M^\ell_{ij}.
\end{equation}
\noindent\textit{where} $M^\ell \in [0,1]^{H^\ell\times W^\ell}$ is the latent-resolution face mask, $C$ is the latent channel count, $\epsilon$ is the sampled noise, and $\hat{\epsilon}_\theta$ is the UNet prediction at $(z_t,t,c)$.

\subsubsection{Verification-Based Identity Loss (Ensemble)}
We encourage the generated face to be verified as the target by several FR models by pushing its similarity to each model's target embedding above a standard threshold.
\begin{equation}
\label{eq:simple_sk}
s_k = \cos\!\big(f_k(\hat{I}),\, e^k_{\mathrm{tgt}}\big),
\end{equation}
\begin{equation}
\label{eq:simple_lid}
\mathcal{L}_{\mathrm{id}} = \frac{1}{K}\sum_{k=1}^{K}\big[(\tau_k+\Delta)-s_k\big]_+ .
\end{equation}
\noindent\textit{where} $f_k(\hat{I})$ is the FR embedding of the generated image under model $\mathcal{F}_k$ (unit-normalized), $e^k_{\mathrm{tgt}}$ is the unit-normalized target embedding, $s_k$ is their cosine similarity, $\tau_k$ is the verification threshold for $\mathcal{F}_k$, $\Delta > 0$ is a small safety margin, and $[x]_+=\max(x,0)$.

\subsubsection{Directional identity (move along source $\to$ target displacement)}
We encourage updates that follow the identity change from the source to the target in FR embedding space, rather than exploiting superficial shortcuts.
\begin{equation}
\label{eq:dir_id}
\begin{aligned}
u_k&=\nu\!\big(f_k(\hat{I})-e^k_{\text{src}}\big), &
v_k&=\nu\!\big(e^k_{\text{tgt}}-e^k_{\text{src}}\big),\\
\mathcal{D}_k&=1-\cos(u_k,v_k), &
\overline{\mathcal{D}}&=\tfrac{1}{K}\sum_{k=1}^K \mathcal{D}_k.
\end{aligned}
\end{equation}
\noindent\textit{where} $f_k(\hat{I})$ is the unit-normalized embedding of the current prediction under FR model $\mathcal{F}_k$; $e^k_{\text{src}}$ and $e^k_{\text{tgt}}$ are the unit-normalized embeddings of source and target; $u_k$ is the \emph{observed} change (prediction minus source), $v_k$ is the \emph{desired} source $\to$ target displacement, and $1-\cos(u_k,v_k)$ penalizes misalignment.

\subsubsection{Source suppression (avoid regress-to-source)}
We discourage the optimizer from drifting back toward the original identity once progress toward the target has been made.
\begin{equation}
\label{eq:source_suppress}
\mathcal{S}_k=\big[\cos\!\big(f_k(\hat{I}),\,e^k_{\text{src}}\big)-m\big]_+,
\qquad
\overline{\mathcal{S}}=\tfrac{1}{K}\sum_{k=1}^K \mathcal{S}_k .
\end{equation}
\noindent\textit{where} $f_k(\hat{I})$ is the unit-normalized embedding of the current prediction under FR model $\mathcal{F}_k$; $e^k_{\text{src}}$ is the unit-normalized source embedding; and $m \in [0,1)$ is a small margin tolerance.

\subsubsection{CLIP-Based Semantic Alignment}
To maintain high-level semantic agreement with the textual prompt while avoiding global drift, we introduce a weak CLIP-guided prior applied exclusively to the facial region:

\begin{equation}
\label{eq:text_align}
\mathcal{L}_{\text{txt}} = 1 - \cos\!\big(\phi_{\text{img}}(\mathrm{crop}(\hat{I},M)),\, \phi_{\text{text}}(c)\big),
\end{equation}

\noindent\textit{where} $c$ denotes the natural-language text prompt conditioning the semantic attributes; $\phi_{\text{img}}$ and $\phi_{\text{text}}$ represent the CLIP-family image and text encoders, respectively; and $\mathrm{crop}(\hat{I},M)$ applies the image-space mask $M$ to ensure the comparison relies solely on facial content. We activate this term only in the later stages of optimization via the schedule $\lambda_{\text{txt}}(i)$, once identity features and denoising have stabilized.

\subsubsection{Optimization Schedules and Timestep Selection}
To ensure stable convergence and prevent early degradation of the facial structure, we employ a stage-wise optimization strategy. The identity-guidance schedule, $s(i)$, is monotonically ramped over the training iterations to smoothly introduce the adversarial shift without destabilizing the latent geometry. Conversely, the semantic regularizer $\lambda_{\text{txt}}(i)$ is implemented as a delayed step function; it remains zero during the initial structural alignment phase and activates only in the final iterations to refine high-level attributes (e.g., hair, accessories). Furthermore, rather than sampling random timesteps as is standard in diffusion training, we anchor the noise level to a fixed diffusion index fraction $\tau$. This stationary noise regime prevents the optimization objective from oscillating and ensures the LoRA updates consistently target the mid-frequency structural features most critical for identity manipulation.
Empirically, we set $\tau = 0.6$; this noise level optimally targets mid-frequency structural facial features for identity morphing, while preserving the low-frequency global composition (e.g., pose and background) necessary for visual stealth.

\subsection{Post-processing: inverse alignment and seamless blending}
\label{sec:post}
Because training operates in a canonical, tightly aligned facial coordinate frame, the edited output must be mapped back to the original image geometry. We first apply the inverse of the alignment affine transform (estimated from facial landmarks) to warp the edited face back to its native pose, scale, and position. This step restores the global structure of the image and ensures spatial consistency with the source context.

However, latent edits may introduce low-frequency tone shifts or local inconsistencies along the face boundary. To mitigate these artifacts, we perform Poisson seamless cloning using the warped prediction as the foreground and the original image as the background. This gradient-domain blending harmonizes illumination, suppresses hard edges, and preserves the background's high-frequency detail while keeping the edited facial region intact. Crucially, we utilize standard Alpha blending or Poisson Normal cloning. This ensures that the generated facial features seamlessly integrate with the background without inadvertently restoring the original identity's gradient structure, as preserving source gradients containing high-frequency textures would reverse the adversarial manipulation and nullify the attack.
In practice, this two-stage post-processing substantially improves visual fidelity by (i) respecting the original scene geometry, (ii) avoiding boundary halos caused by the VAE decoder, and (iii) ensuring that the final output remains photorealistic under arbitrary lighting and backgrounds.

\section{Experiments}
\label{sec:experiments}

In this work, we operate under a realistic transfer-based black-box setting for the final victim system, while leveraging a white-box setting during the optimization of our surrogate ensemble.  Specifically, we assume the attacker does not have access to the target system. Instead, we utilize the gradients of a surrogate ensemble consisting of three other models to optimize the adversarial features. The potency of Adv-TGD relies on the cross-model transferability of these latent-space perturbations. We evaluate the effectiveness of our method in \emph{black-box} attack settings against robust face recognition (FR) models and compare with recent state-of-the-art. We also report perceptual fidelity metrics and ablations of key components.

\begin{table*}[t]
\centering
\caption{Attack Success Rate (ASR\%) on \textbf{FFHQ} and \textbf{CelebA-HQ}}
\label{tab:asr_results_columnar}
\resizebox{\linewidth}{!}{%
\begin{tabular}{ll|cccc|cccc|c}
\toprule
& & \multicolumn{4}{c|}{\textbf{FFHQ}} & \multicolumn{4}{c|}{\textbf{CelebA-HQ}} & \\
\cmidrule(lr){3-6} \cmidrule(lr){7-10}
\textbf{Category} & \textbf{Method} & \textbf{IR152} & \textbf{IRSE50} & \textbf{FaceNet} & \textbf{Mobile} & \textbf{IR152} & \textbf{IRSE50} & \textbf{FaceNet} & \textbf{Mobile} & \textbf{Avg.} \\
\midrule
& Clean & 3.20 & 2.20 & 2.10 & 4.90 & 2.10 & 5.61 & 0.80 & 13.60 & 4.31 \\
\midrule
\multirow{3}{*}{Noise-Based} 
& TI\mbox{-}DIM~\cite{dong2019evading} & 42.26 & 63.91 & 14.28 & 51.71 & 34.08 & 60.51 & 13.27 & 51.25 & 41.41 \\
& TIP\mbox{-}IM~\cite{yang2021towards} & 44.86 & 65.36 & 58.03 & 50.47 & 40.02 & 55.57 & 37.90 & 48.07 & 50.04 \\
& P3\mbox{-}Mask~\cite{chow2024personalized} & 70.98 & 82.79 & 56.18 & 68.57 & 71.27 & 80.90 & 58.43 & 67.55 & 69.58 \\
\midrule
\multirow{4}{*}{\shortstack[l]{Makeup /\\Attribute}} 
& CLIP2Protect~\cite{shamshad2023clip2protect} & 50.56 & 83.93 & 43.66 & 74.00 & 46.20 & 78.53 & 41.29 & 71.43 & 61.20 \\
& DFPP~\cite{shamshad2024makeup} & 52.62 & 87.91 & 50.57 & 77.69 & 45.00 & 78.37 & 44.01 & 69.97 & 63.24 \\
& DiffAM~\cite{sun2024diffam} & 65.22 & 87.59 & 63.01 & 86.87 & 62.14 & 86.97 & 61.10 & 81.97 & 74.49 \\
\midrule
\multirow{4}{*}{\shortstack[l]{Portrait /\\Semantic}} 
& Adv\mbox{-}Diffusion~\cite{liu2024adv} & 49.40 & 79.31 & 29.91 & 65.49 & 51.25 & 79.22 & 33.90 & 68.66 & 57.14 \\
& Adv-CPG ~\cite{wang2025adv} & 75.26 & 91.03 & 62.84 & 89.94 & 76.96 & 88.72 & \underline{63.50} & 87.95 & 79.65 \\
& TCA$^2$~\cite{li2025transferable} & 53.63 & 77.31 & 41.62 & 72.21 & 52.50 & 76.21 & 40.72 & 71.92 & 60.76 \\
& DiffAIM~\cite{wang2025diffusion} & \underline{82.20} & \underline{91.04} & \textbf{68.21} & \underline{94.21} & \underline{80.56} & \underline{90.37} & \textbf{64.14} & \underline{93.10} & \underline{82.85} \\
& \textbf{Adv-TGD (Ours)} 
& \textbf{88.80} & \textbf{94.60} & \underline{63.40} & \textbf{95.40} & \textbf{90.60} & \textbf{94.80} & 63.40 & \textbf{96.20} & \textbf{85.90} \\
\bottomrule
\end{tabular} }
\end{table*}

\subsection{Experimental Setting}

\subsubsection{Dataset}
We evaluate Adv-TGD using two widely adopted, high-quality facial image datasets: CelebA-HQ~\cite{karras2017progressive} and FFHQ~\cite{karras2019style}. In line with the evaluation protocols described in~\cite{wang2025adv, liu2024adv}, we randomly select a subset of 1,000 source images representing distinct identities from each dataset, along with an additional 5 distinct images to serve as target identities. To systematically evaluate the attack, we divide the 1,000 source images into five mutually exclusive groups of 200 images. Each group is then exclusively assigned to one of the 5 target identities. Consequently, this yields a total of 1,000 unique source--target image pairs per dataset, distributed evenly across five distinct target identities.


\subsubsection{Implementation details}
We build on Stable Diffusion v2.1 at $768{\times}768$. For each $(I_{\text{src}}, I_{\text{tgt}})$ pair, we fine-tune a \emph{pair-specific} LoRA adapter injected into U-Net cross-attention projections \texttt{to\_q}, \texttt{to\_k}, \texttt{to\_v}, \texttt{to\_out.0}. We use LoRA rank $r=16$ and $\alpha=64$ with dropout $0.08$. Training runs for $65$ steps (AdamW, base LR $3{\times}10^{-5}$ with late linear decay to $1{\times}10^{-5}$). We use a single denoising slice per step with a fixed time index fraction $\tau=0.6$, DDPM scheduler, and latent blending under a soft face mask. OpenCLIP text guidance is applied only in the late phase, and perceptual/TV terms are weak and masked to the face region. Re-evaluation scores are computed after seamless blending back into the original frame.

\subsubsection{Evaluation Metrics}
We evaluate our approach by independently assessing attack performance and image quality. To measure attack performance, we adopt the Attack Success Rate (ASR)~\cite{jia2022adv, liu2024adv, wang2025adv}, which quantifies the proportion of adversarial examples that successfully deceive the target model:
\begin{equation}
\text{ASR} = \frac{1}{N} \sum_{i=1}^{N} \mathbb{I}\Big( \cos\big(\mathcal{F}_{\text{eval}}(I_{\text{tgt}}^{(i)}), \mathcal{F}_{\text{eval}}(I_{\text{adv}}^{(i)})\big) > \gamma \Big) \times 100\%,
\end{equation}
\noindent\textit{where} $\mathbb{I}(\cdot)$ denotes the indicator function, $N$ represents the total number of face images, and $\gamma$ is the decision threshold. $I_{\text{tgt}}^{(i)}$ and $I_{\text{adv}}^{(i)}$ correspond to the $i$-th target and adversarial face images, respectively. $\mathcal{F}_{\text{eval}}(\cdot)$ is the feature extraction network of the victim model. The threshold $\gamma$ is set to yield a 0.01 False Acceptance Rate (FAR) for each victim model, following the same configuration as~\cite{jia2022adv, liu2024adv, wang2025adv}. For image quality assessment, we employ three standard metrics: Fréchet Inception Distance (FID), Peak Signal-to-Noise Ratio (PSNR), and Structural Similarity Index (SSIM).

\subsection{Comparison with SOTA Methods}

\subsubsection{Quantitative results}
Table~\ref{tab:asr_results_columnar} summarizes the attack success rates (ASR) under the black-box setting; underlined entries denote the highest value in each column. For face verification, we set the decision threshold $\gamma$ per model at $\text{FAR}=0.01$. Our method achieves strong black-box transferability, particularly on IR152, IRSE50, and MobileFace, and substantially outperforms representative noise-, makeup-, and semantic-based baselines. To further validate the imperceptibility of the proposed adversarial attack, we evaluate the visual quality of adversarial examples compared with representative SOTA baselines. Following common practice, we adopt Frechet Inception Distance (FID)~\cite{heusel2017gans}, Peak Signal-to-Noise Ratio (PSNR), and Structural Similarity Index (SSIM) as evaluation metrics. Table~\ref{tab:quality_results} reports the quantitative results aggregated across multiple runs. While Adv-Diffusion and DiffAIM achieve lower FID scores, our method achieves the highest SSIM (0.981) and high PSNR (28.18 dB). This indicates that while baseline methods may match the unconditional image distribution slightly better, \textbf{Adv-TGD} significantly outperforms them in pixel-level structural preservation. For stealthy adversarial impersonation, this structural and spatial consistency is arguably more critical to ensure seamless facial blending and minimal visual distortion.

\subsubsection{Image quality assessment}
To evaluate the visual imperceptibility of the proposed adversarial generation framework, we further analyze the perceptual quality of generated images. In general, lower FID values indicate that the generated images are closer to the real image distribution, while higher PSNR and SSIM values correspond to better reconstruction fidelity and structural consistency. In addition to quantitative evaluation, we also present qualitative comparisons in Fig.~\ref{fig:teaser}. The results show that Adv-TGD produces visually coherent faces that preserve lighting, expression, and background consistency while successfully aligning with the target identity. Compared with existing approaches, our method introduces fewer visual artifacts and maintains more natural facial appearance, highlighting the effectiveness of the proposed semantic-guided adversarial generation framework.

\begin{table}[t]
\centering
\caption{Perceptual quality comparison. Lower is better for FID; higher is better for PSNR/SSIM.}
\label{tab:quality_results}
\renewcommand{\arraystretch}{1.1} 
\resizebox{0.65\linewidth}{!}{%
\begin{tabular}{lccc}
\toprule
\textbf{Method} & \textbf{FID}$\downarrow$ & \textbf{PSNR}$\uparrow$ & \textbf{SSIM}$\uparrow$ \\
\midrule
TIP-IM & 38.73 & \textbf{33.21} & \underline{0.924} \\
Adv-Diffusion & \textbf{22.58} & 27.85 & 0.805 \\
DiffAM & 26.10 & 20.53 & 0.886 \\
Adv-CPG & 26.07 & \underline{28.23} & 0.897 \\
DiffAIM & \underline{23.23} & 24.39 & 0.739 \\
\textbf{Adv-TGD (Ours)} & 25.90 & 28.18 & \textbf{0.981} \\
\bottomrule
\end{tabular}%
}
\end{table}

\subsubsection{Evaluation on commercial FR API}
\begin{figure*}[ht!]
  \centering
  \includegraphics[width=0.85\linewidth]{ 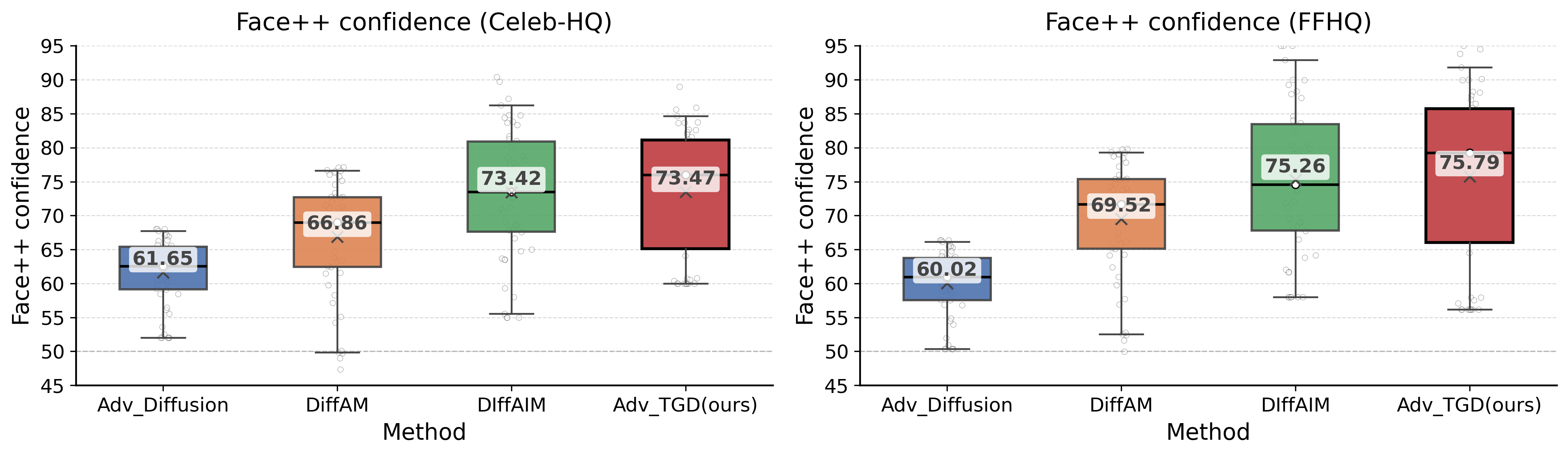}
  \caption{Face++ confidence scores ($\uparrow$) returned from the commercial API for four attack methods on \textbf{CelebA-HQ} (left) and \textbf{FFHQ} (right).}
  \label{fig:facepp_boxplots}
\end{figure*}

To assess black-box transferability under a real-world deployment scenario, we further evaluate all methods on a production-grade commercial face verification service. For each dataset (CelebA-HQ and FFHQ), we sample 100 source--target identity pairs and generate an adversarial source image for every method. Each adversarial image is then submitted to the Face++ \emph{Compare} API together with its corresponding target image. The API returns a proprietary confidence score in $[0,100]$ that reflects the similarity between the two inputs, where larger values indicate stronger impersonation. Visualization of the empirical distributions is provided in Fig.~\ref{fig:facepp_boxplots}. 

\section{Ablation Study and Mechanism Analysis}
\label{sec:ablation_study}

To rigorously evaluate the contribution of each component in our \textbf{Adv-TGD} framework, we conduct a comprehensive ablation study using a leave-one-out strategy. Experiments are performed on a randomly sampled subset of the CelebA-HQ dataset using the industry-standard \textbf{IR152} recognizer at \textbf{FAR=0.01}. The results in Table~\ref{tab:full_ablation} highlight the trade-offs between identity manipulation effectiveness (Attack Success Rate -- ASR), visual fidelity (PSNR, SSIM), and computational efficiency.
\begin{table*}[t]
\centering
\caption{\textbf{Ablation analysis of Adv-TGD on CelebA-HQ.} We evaluate different masking strategies, loss components, and blending techniques. Runtime is measured as the average time to generate one adversarial example on a single NVIDIA A100 GPU. Our proposed \textbf{SGSM} achieves the best balance between adversarial effectiveness and visual fidelity.}
\label{tab:full_ablation}
\resizebox{\linewidth}{!}{%
\begin{tabular}{l|cccc|l}
\toprule
\textbf{Variant} & \textbf{ASR (\%)} $\uparrow$ & \textbf{PSNR} $\uparrow$ & \textbf{SSIM} $\uparrow$ & \textbf{Runtime (s)} $\downarrow$ & \textbf{Observation} \\
\midrule
\multicolumn{6}{c}{\textit{Masking Strategies}} \\
\midrule
Saliency Mask & 77.50 & \textbf{28.83} & \textbf{0.984} & 85.2 & High fidelity but weaker identity manipulation. \\
Full Image Editing & 88.50 & 23.54 & 0.958 & 42.4 & Strong attack but severe background distortion. \\
Geometric Mask & 87.50 & 25.29 & 0.968 & \textbf{36.7} & Landmark constraint improves structural edits. \\

\textbf{SGSM (Ours)} & \textbf{91.00} & 28.18 & 0.979 & 88.1 & \textbf{Best attack–quality trade-off.} \\
\midrule
\multicolumn{6}{c}{\textit{Loss Components}} \\
\midrule
w/o Directional Loss ($L_{dir}$) & 86.20 & 26.41 & 0.985 & 84.9 & Less consistent identity shift. \\
w/o Source Suppression ($L_{src}$) & 84.60 & 26.38 & 0.984 & 85.1 & Source identity partially preserved. \\
w/o Text Guidance ($L_{txt}$) & 87.80 & 26.47 & 0.984 & 83.5 & Slight drop in semantic consistency. \\
\midrule
\multicolumn{6}{c}{\textit{Blending Strategy}} \\
\midrule
Laplacian Pyramid Blending & 90.00 & 27.18 & 0.977 & 112.4 & Similar performance but higher cost. \\
\bottomrule
\end{tabular}
}
\end{table*}

\subsection{Impact of Masking Strategy}
The masking strategy is the most influential factor in our pipeline. The \textbf{Saliency-Only Mask} baseline utilizes only the FR model's attention maps; while it preserves high visual quality, it often fails to cover identity-leaking features that the model deems non-salient (\textit{e.g.}, specific jawline structures), resulting in a lower ASR of 77.5\%. \textbf{No Masking} (Full Image Editing) yields a high ASR (88.5\%) but drastically degrades fidelity by destroying background consistency. The \textbf{Parsing Mask} (Geometric Mask) utilizes a landmark-based convex hull to strictly mask the anatomical face, achieving a strong ASR of 87.5\%; this confirms that geometric constraints force the optimizer to make structural changes rather than relying on superficial background noise. Our proposed \textbf{SGSM} achieves the optimal \textbf{ASR of 91.0\%} by fusing anatomical landmarks with target-aware gradients, ensuring the optimizer targets specific identity-bearing ``hotspots'' while maintaining superior fidelity.

\subsection{Impact of Loss Components and Text Guidance}
Removing the \textbf{Directional Loss} ($L_{dir}$) lowers the mean cosine similarity, validating its role in steering the identity shift along the semantic vector in embedding space rather than just pushing it away randomly. When using only a basic hinge loss without the directional term, we observe that some adversarial examples partially drift between source and target identities, especially under challenging poses. In the absence of \textbf{Source Suppression} ($L_{src}$), the optimizer tends to ``snap back'' to source identity features, highlighting $L_{src}$ as a critical anchor. \textbf{Text Guidance} ($L_{txt}$) acts as a semantic regularizer; removing it results in a slight decrease in semantic consistency, confirming its effectiveness as a lightweight refinement once identity and denoising have stabilized. Furthermore, applying strong text guidance from the beginning may overemphasize textual attributes at the expense of identity stability, demonstrating why our late-stage application is necessary.

\subsection{Impact of Blending Techniques}
\textbf{Laplacian Pyramid Blending} performed similarly to our final model (90.0\% ASR) but is computationally more expensive. Thus, we utilize Gaussian-feathered alpha blending as it provides the optimal balance of efficiency and quality.

\subsection{Visual and Spectral Analysis}
To investigate the mechanism of the identity shift, we perform a spatial and frequency-domain analysis of the generated adversarial perturbations. 

\subsubsection{Attention Disruption and Dispersion} Fig.~\ref{fig:mechanism_analysis}(b-d) visualizes the FR model's attention maps. While the source image triggers concentrated activations tightly clustered around identity-critical landmarks (\textit{e.g.}, ocular and nasal regions), \textbf{Adv-TGD} successfully neutralizes the recognizer by dispersing its spatial attention. We quantify this effect by measuring the spatial spread (RMS variance) and informational entropy. In notable instances, our method can expand the spatial spread of the model's attention by up to \textbf{8.4\%} while increasing spatial entropy by as much as \textbf{1.5\%}. As shown in the difference map (Fig.~\ref{fig:mechanism_analysis}d), the attack does not simply add noise but systematically dismantles the focal points required for identification.

\subsubsection{Frequency Domain Analysis} 
\begin{figure}[tb]
    \centering
    \includegraphics[width=0.85\linewidth]{ 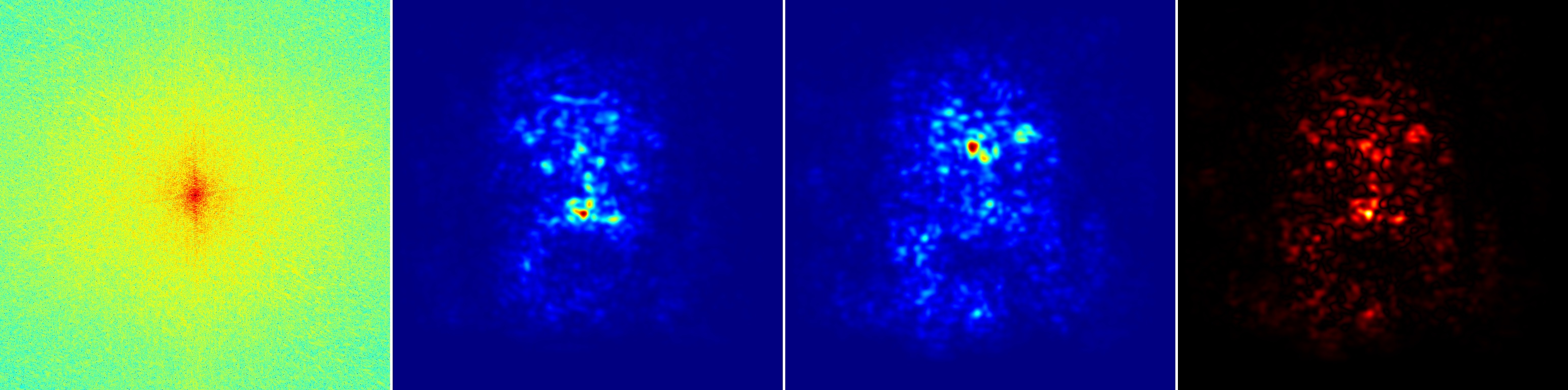} 
    \caption{\textbf{Mechanism of Identity Neutralization.} From left to right: (a) spectral analysis confirming the attack targets structural frequencies. (b-c) visualization of FR focal attention dispersal. (d) saliency difference map $|S_{adv} - S_{src}|$, revealing the precise anatomical regions neutralized by \textbf{Adv-TGD}.}
    \label{fig:mechanism_analysis}
\end{figure}

Fig.~\ref{fig:mechanism_analysis}(a) displays the Discrete Fourier Transform (DFT) of the perturbation. The energy is heavily concentrated in the low-to-mid frequency bands (central region), indicating that the attack targets semantic facial structures, such as jawline geometry and facial depth, rather than relying on high-frequency pixel noise.

\subsection{Optimization Dynamics}
\begin{figure}[t!]
    \centering
    \includegraphics[width=0.85\linewidth]{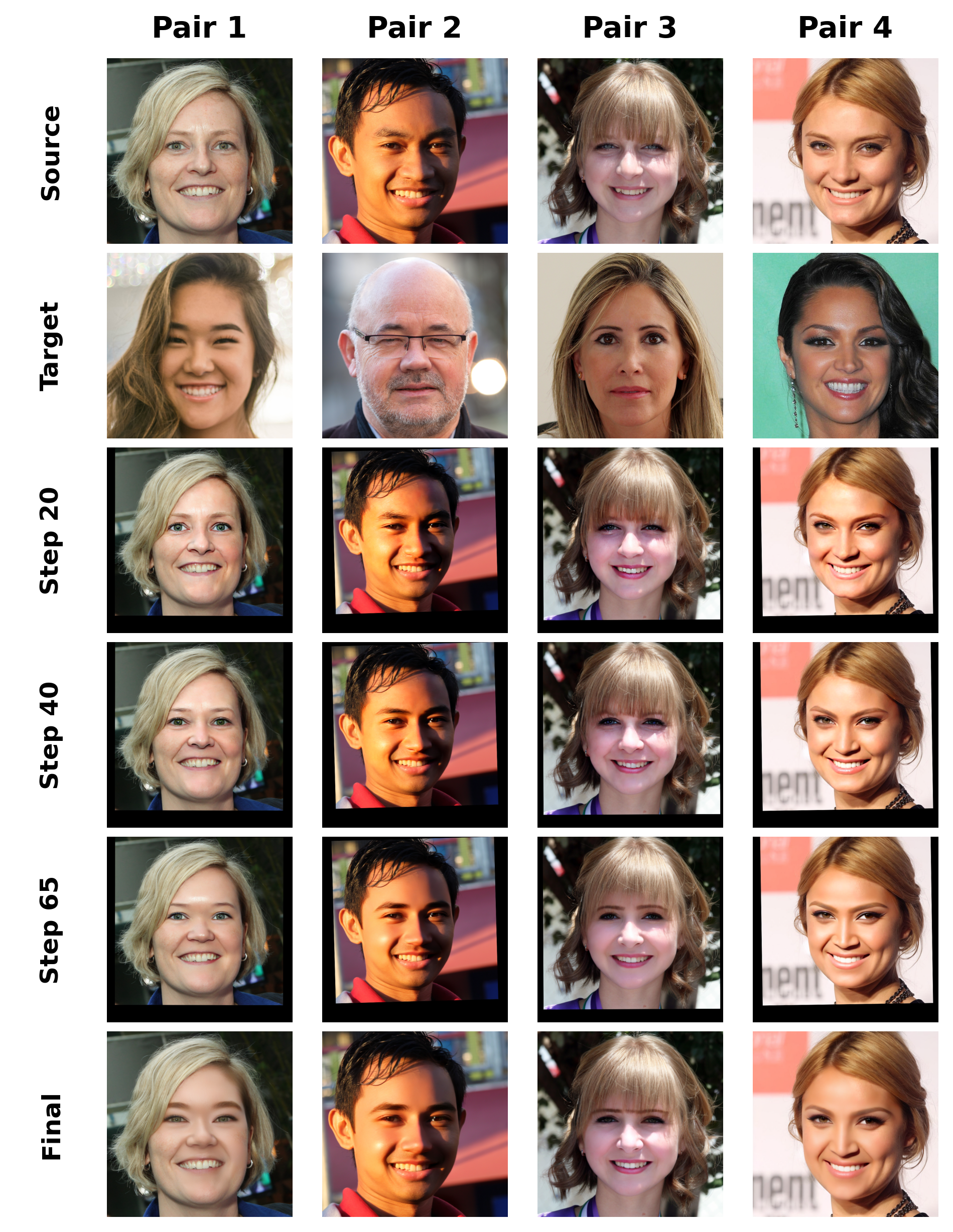} 
    \caption{\textbf{Evolution of the Adversarial Identity.} Visualization of the Adv-TGD optimization process across different steps. The source face gradually adopts the structural and semantic features of the target identity over iterations, culminating in a seamlessly blended final output.}
    \label{fig:opt_progression}
\end{figure}

To further understand the generative process, we visualize the evolution of the adversarial face during the per-pair LoRA optimization. Fig.~\ref{fig:opt_progression} illustrates the progression across 65 optimization steps for multiple source--target pairs. In the early stages (e.g., Step 20), the optimizer begins altering the foundational facial geometry and low-frequency depth cues. As training progresses, fine-grained identity features such as the jawline, eye shape, and skin textures are aggressively refined to match the target. The \textit{Final} row demonstrates the result of our post-processing pipeline, where the optimized adversarial face is seamlessly blended back into the original source lighting and background context, yielding a high-fidelity, adversarial image.

\section{Experimental Prompts}
\label{sec:prompts_visuals}

\begin{figure}[htbp]
    \centering
    \includegraphics[width=0.6\linewidth]{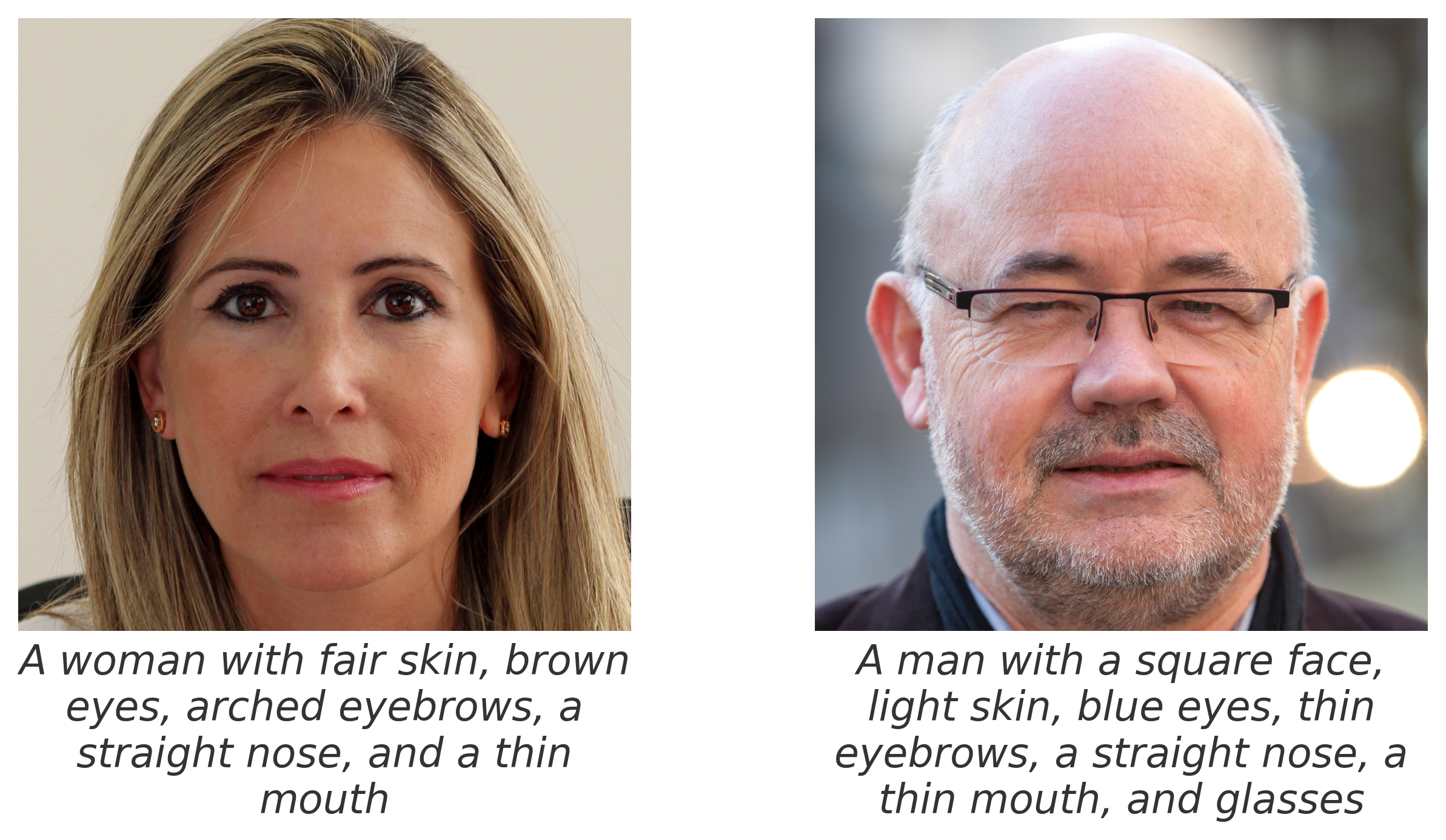}
    \caption{Example texts used during the late-stage guidance phase.}
    \label{fig:supp_examples}
\end{figure}

Fig.~\ref{fig:supp_examples} displays example text used to generate the adversarial examples shown in experiments. These texts are descriptive yet broad enough to allow the optimization to dictate facial structure. The specific prompt used to generate these textual descriptions is: \textit{``Provide a detailed description of the person's facial features.''}

\section{Extensions and Generalization}
\label{sec:extensions}

\subsection{Evaluation on LADN Dataset}
\label{sec:ladn_results}
To further verify the generalization capability and robustness of \textbf{Adv-TGD}, we conducted additional experiments on the LADN dataset~\cite{gu2019ladn}. Unlike the aligned CelebA-HQ dataset, LADN contains images with significant variations in makeup, lighting, and pose, presenting a more challenging ``in-the-wild'' scenario. As reported in Table~\ref{tab:asr_ladn}, \textbf{Adv-TGD} achieves consistently strong attack performance across all evaluated face recognition models under these challenging conditions.

\begin{table}[h!]
    \centering
    \caption{Attack Success Rate (ASR\%) on the \textbf{LADN} dataset. Comparison against state-of-the-art methods.}
    \label{tab:asr_ladn}
    \resizebox{0.75\linewidth}{!}{%
    \begin{tabular}{l|cccc}
        \toprule
        \textbf{Method} & \textbf{IR152} & \textbf{IRSE50} & \textbf{FaceNet} & \textbf{MobileFace} \\
        \midrule
        Clean & 3.71 & 2.79 & 0.60 & 5.12 \\
        \midrule
        Clip2Protect & 54.31 & 88.57 & 49.91 & 78.94 \\
        DiffAM & 66.68 & 91.26 & 66.16 & 88.68 \\
        DiffAIM & 74.21 & 92.67 & \textbf{68.44} & 83.45 \\
        \midrule
        \textbf{Adv-TGD (Ours)} & \textbf{90.80} & \textbf{94.30} & 65.71 & \textbf{92.80} \\
        \bottomrule
    \end{tabular}
    }
\end{table}

\subsection{Extension to General Object Classification}
\label{sec:non_face}

To demonstrate the universality of the proposed \textbf{Adv-TGD} framework beyond face recognition, we extended our approach to attack general image classification models.

\subsubsection{Experimental Setup}
For this extension, we adapted the pipeline to target a pre-trained \textbf{ResNet-50} classifier. Instead of SGSM, we utilized class-activation maps (CAM) to derive the saliency signal, allowing the diffusion model to modify global semantic features relevant to the target class while preserving the structural essence of the source object. Text guidance was adjusted to describe the target ImageNet category (\textit{e.g.}, ``A photo of a Golden Retriever'').

\subsubsection{Qualitative Results}
\begin{figure}[ht!]
    \centering
    \includegraphics[width=0.6\linewidth]{ 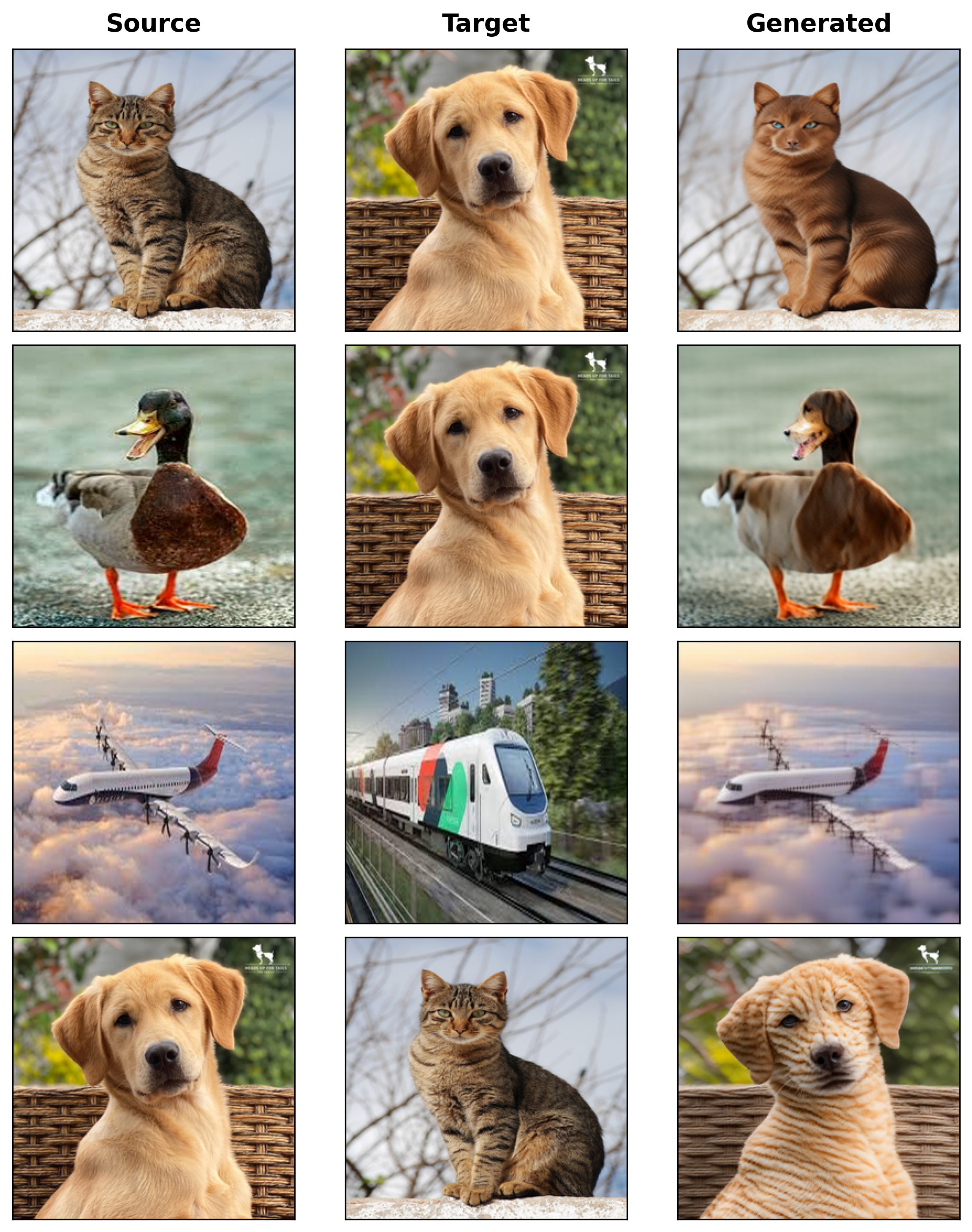} 
    \caption{\textbf{General Object Classification Attack.} Qualitative results of \textbf{Adv-TGD} attacking a ResNet-50 classifier on ImageNet.}
    \label{fig:non_face_results}
\end{figure}

Fig.~\ref{fig:non_face_results} illustrates the adversarial examples generated by \textbf{Adv-TGD} on ImageNet samples. As observed, the framework successfully injects subtle semantic features of the target class while maintaining the overall spatial structure of the source object.

\subsection{Generalization Across Architectures}
\label{sec:generalization}

A core strength of the \textbf{Adv-TGD} framework is its architectural agnosticism. While our primary experiments utilize Stable Diffusion 2.1, the underlying mechanism is transferable to diverse generative and discriminative backbones.

\subsubsection{Stable Diffusion 1.5 (SD 1.5)}
\label{sec:gen_sd15}
Stable Diffusion 1.5 relies on a similar UNet-based architecture but utilizes a different text encoder. The transition is seamless, and the \textbf{SGSM} strategy remains identical. We observed that Adv-TGD achieves similar or slightly higher ASR on SD 1.5 compared to SD 2.1, suggesting the vulnerability is inherent to continuous latent representations of identity. Quantitative results for different masking strategies on SD 1.5 are provided in Table~\ref{tab:sd15_results}.

\begin{table}[h!]
    \centering
    \caption{Performance of Adv-TGD masking variants on Stable Diffusion 1.5.}
    \label{tab:sd15_results}
    \resizebox{0.75\linewidth}{!}{%
    \begin{tabular}{lccc}
        \toprule
        \textbf{Variant} & \textbf{ASR (\%)} $\uparrow$ & \textbf{PSNR} $\uparrow$ & \textbf{SSIM} $\uparrow$ \\
        \midrule
        Saliency Mask & 69.00 & 17.63 & 0.796 \\
        Full Image Editing & 84.00 & 16.42 & 0.764 \\
        Geometric Mask (Parsing) & 86.80 & 16.52 & 0.789 \\
        \textbf{SGSM (Ours)} & 87.50 & 18.71 & 0.819 \\
        \bottomrule
    \end{tabular}
    }
\end{table}


\begin{figure*}[t!]
    \centering
    \includegraphics[width=0.95\linewidth]{ 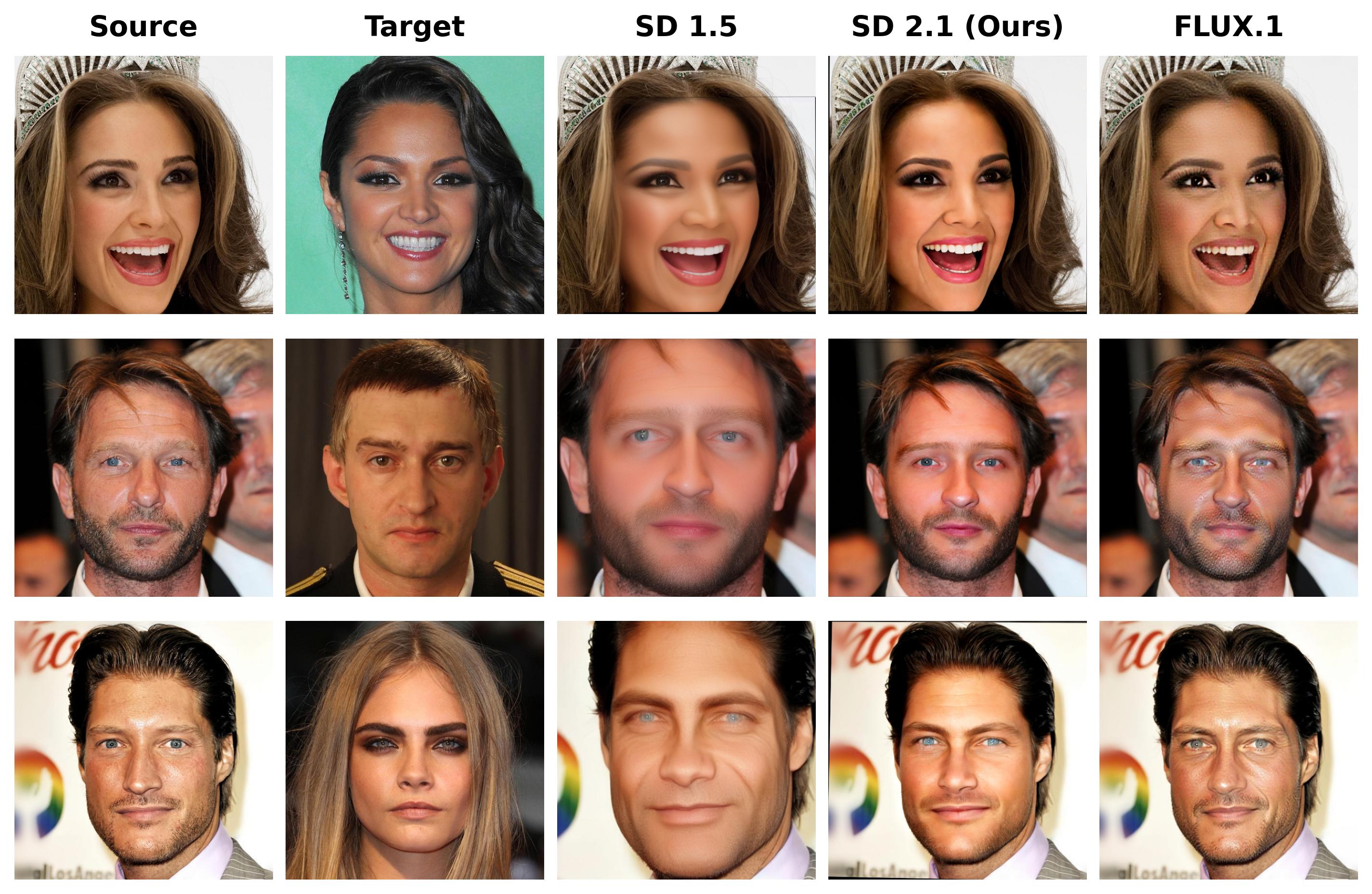} 
    \caption{\textbf{Cross-Architecture Generalization.} Visual comparison of Adv-TGD generated adversarial faces across different generative backbones. From left to right: Source, Target, Stable Diffusion 1.5, Stable Diffusion 2.1 (Primary), and FLUX.1. The framework consistently achieves robust identity transfer and structural preservation.}
    \label{fig:cross_arch}
\end{figure*}

\subsubsection{Flux.1 (Transformer-Based Flow Matching)}
\label{sec:gen_flux}

To evaluate the architectural agnosticism of \textbf{Adv-TGD}, we adapted the framework to the state-of-the-art \textbf{FLUX.1-dev} model. This transition required a fundamental shift in the optimization objective, latent handling, and noise regularization to accommodate a non-UNet architecture operating via rectified flow:

\begin{itemize}
    \item \textbf{Rectified Flow Objective:} Unlike the epsilon-prediction used in our primary experiments, FLUX utilizes a flow-matching formulation. We reformulated our MSE loss to target the velocity vector $v$, where the transformer predicts the linear trajectory between the source latent and random noise: $\mathcal{L}_{mse} = \| \hat{v}_\theta - (\epsilon - z_{\text{src}}) \|^2$. This rectified flow training provides significantly higher gradient stability during the identity shift.
    
    \item \textbf{Dynamic Timestep Sampling:} Continuous flow-matching models are highly susceptible to ``latent burn-in'' if gradients are backpropagated through a fixed noise level. To ensure the adversarial identity features are robust and generalize across the flow trajectory, we replaced the fixed timestep anchor with dynamic sampling $t \sim \mathcal{U}(0.2, 0.5)$. This forces the transformer to learn structural semantic shifts rather than exploiting a specific noise manifold.
    
    \item \textbf{Sequence-Spatial Mapping and Pixel-Space Masking:} Since the \textbf{Multimodal Diffusion Transformer (MMDiT)} operates on flattened $2\times2$ patch sequences, we implemented differentiable \texttt{pack} and \texttt{unpack} operations. Crucially, we found that applying our facial mask in the latent space introduces severe VAE decoding artifacts in FLUX. Therefore, we extract the flow predictions, fully decode them to the spatial pixel domain, and apply a strict landmark-bounded Parsing Mask (Method F) \textit{before} FR feature extraction.
    
    \item \textbf{High-Frequency Artifact Suppression:} Massive transformer backbones have a tendency to achieve adversarial success by injecting high-frequency static rather than purely structural identity changes, which can degrade visual fidelity. While our dual-constraint perceptual regularization ($\mathcal{L}_{reg} = \lambda_{tv}\mathcal{L}_{TV} + \lambda_{lpips}\mathcal{L}_{LPIPS}$) helps suppress this, we observe that transformer-based flow matching still exhibits a higher baseline of pixel-level noise compared to UNet architectures.
    
    \item \textbf{Transformer LoRA Injection:} We localized the trainable parameters to the transformer's cross-attention projections (\texttt{to\_q, to\_k, to\_v, to\_out.0}). While this enables strong semantic identity manipulation, the resulting gradients in the transformer space tend to produce rougher skin textures compared to the smooth synthesis of SD 2.1.
\end{itemize}

As shown in Table~\ref{tab:flux_results}, adapting the framework to FLUX.1 successfully yields strong attack transferability while capitalizing on the transformer backbone to achieve high visual fidelity (PSNR $\sim$29 dB, SSIM $>0.98$). A core strength of the \textbf{Adv-TGD} framework is its architectural agnosticism. While our primary experiments utilize Stable Diffusion 2.1, the underlying mechanism is transferable to diverse generative backbones. As illustrated in Fig.~\ref{fig:cross_arch}, Adv-TGD produces consistent adversarial identity transformations across multiple architectures, including UNet-based diffusion models (Stable Diffusion 1.5 and 2.1) and transformer-based generative models (FLUX.1).

\begin{table}[h!]
    \centering
    \caption{Performance of Adv-TGD masking variants on the FLUX.1 architecture.}
    \label{tab:flux_results}
    \resizebox{0.75\linewidth}{!}{%
    \begin{tabular}{lccc}
        \toprule
        \textbf{Variant} & \textbf{ASR (\%)} $\uparrow$ & \textbf{PSNR} $\uparrow$ & \textbf{SSIM} $\uparrow$ \\
        \midrule
        Saliency Mask & 66.50 & 29.85 & 0.986 \\
        Full Image Editing & 84.00 & 26.40 & 0.955 \\
        Geometric Mask (Parsing) & 79.00 & 29.15 & 0.980 \\
        \textbf{SGSM (Ours)} & 83.50 & 29.30 & 0.982 \\
        \bottomrule
    \end{tabular}
    }
\end{table}

\section{Conclusion}
\label{sec:conclusion}
We introduced \textit{Adv-TGD}, a text-guided, per-pair LoRA framework for adversarial face synthesis built on a frozen Stable Diffusion 2.1 backbone. By introducing \textit{Salience-Guided Semantic Masking (SGSM)} and a unified identity objective comprising ensemble hinges, directional guidance, and source suppression\textit{Adv-TGD} enables targeted identity manipulation that is both parameter-efficient and structurally consistent. Unlike traditional adversarial noise, our method targets low-to-mid frequency facial features to neutralize recognizers by dispersing their spatial attention. Extensive experiments on CelebA-HQ and FFHQ under a strict black-box setting demonstrate that \textit{Adv-TGD} achieves exceptional transferability across industry-standard FR models. Our method attains a state-of-the-art mean ASR of \textbf{85.90\%}, outperforming representative noise-based, makeup-based, and semantic baselines. Crucially, \textit{Adv-TGD} maintains superior pixel-level perceptual quality, achieving a PSNR of 28.18 dB and an SSIM of 0.981, ensuring that the adversarial faces remain photorealistic and structurally faithful to the source context. Our results reveal a previously underexplored vulnerability of modern face recognition systems to generative diffusion-based attacks. By combining semantic guidance and localized identity manipulation, Adv-TGD achieves strong impersonation success while preserving high perceptual realism. These findings highlight the need for more robust biometric systems capable of defending against generative adversarial attacks. Beyond academic face recognition benchmarks, we demonstrated the framework's broad universality and architectural agnosticism. Evaluations on commercial systems (Face++), in-the-wild datasets (LADN), and general object classification (ImageNet) confirm its robust real-world applicability. Furthermore, cross-architecture validations on Stable Diffusion 1.5 and the state-of-the-art transformer-based FLUX.1 model highlight that the fusion of semantic constraints with target-aware saliency is highly adaptable. Overall, our results demonstrate that text-guided localized generative adaptation offers an explainable, practical, and highly effective paradigm for adversarial impersonation against unauthorized recognition.


\bibliographystyle{plainnat}
\bibliography{references}

\section{Ethical Considerations}
Our research explores vulnerabilities in face recognition systems using generative AI. While the proposed Adv-TGD framework can synthesize impersonation images, we emphasize that this work is intended for defensive security analysis and privacy protection. All experiments were conducted using publicly available research datasets (CelebA-HQ, FFHQ, LADN). We do not release any pre-trained LoRA weights for specific individuals, and our methodology is designed to inform the development of more robust biometric verification systems against generative threats.




\end{document}